\pdfoutput=1

\documentclass[11pt]{article}
\usepackage[table,xcdraw]{xcolor}

\usepackage[preprint]{acl}

\usepackage{times}
\usepackage{latexsym}
\usepackage{amssymb}

\usepackage[T1]{fontenc}

\usepackage[utf8]{inputenc}

\usepackage{microtype}

\usepackage{inconsolata}

\usepackage{graphicx}
\usepackage{booktabs}
\usepackage{amsmath}
\usepackage{xcolor}
\usepackage{algorithm2e}
\usepackage{enumitem}
\usepackage{listings}
\usepackage{multirow}

\title{EditGRPO: Reinforcement Learning with Post‑Rollout Edits for Clinically Accurate Chest X‑Ray Report Generation}

\author{Kai Zhang\textsuperscript{1,2}\thanks{Work done as an intern at NEC Laboratories America.}, 
Christopher Malon\textsuperscript{1}, Lichao Sun\textsuperscript{2}, Martin Renqiang Min\textsuperscript{1} \\
\textsuperscript{1}NEC Laboratories America, \textsuperscript{2}Lehigh University\\
\texttt{kaz321@lehigh.edu, malon@nec-labs.com, lis221@lehigh.edu, renqiang@nec-labs.com}\\
Code available at: \url{https://github.com/taokz/EditGRPO}
}

\begin{document}
\maketitle
\begin{abstract}
Radiology report generation requires advanced medical image analysis, effective temporal reasoning, and accurate text generation. Although recent innovations, particularly multimodal large language models, have shown improved performance, their supervised fine-tuning (SFT) objective is not explicitly aligned with clinical efficacy. In this work, we introduce \textbf{EditGRPO}, a mixed-policy reinforcement learning algorithm designed specifically to optimize the generation through clinically motivated rewards. EditGRPO integrates on-policy exploration with off-policy guidance by injecting sentence-level detailed corrections during training rollouts. This mixed-policy approach addresses the exploration dilemma and sampling efficiency issues typically encountered in RL. Applied to a Qwen2.5-VL-3B, EditGRPO outperforms both SFT and vanilla GRPO baselines, achieving an average improvement of 3.4\% in clinical metrics across four major datasets. Notably, EditGRPO also demonstrates superior out-of-domain generalization, with an average performance gain of 5.9\% on unseen datasets. 
\end{abstract}

\begin{figure*}
    \centering
    \includegraphics[width=\linewidth]{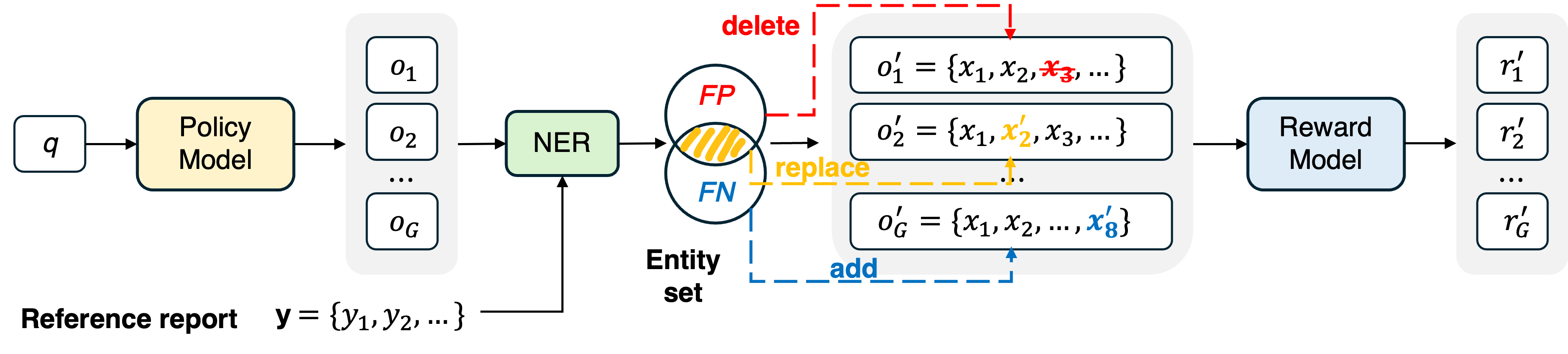}
    \caption{The graphical diagram illustrates the \textit{post-rollout-edit} technique used in the proposed EditGRPO algorithm. For each rollout, the generated response $o$ is edited based on the gold-standard or reference report $\mathbf{y}$ at the sentence level. This includes replacing incorrect or false positive (FP) sentences $x$. For example, if the reference contains ``cardiomegaly'' but the generated report states ``the heart is within normal limits,'' the incorrect sentence is replaced. Additionally, missing findings, referred to as false negatives (FN), can be added based on the reference report.}
    \label{fig:overview}
\end{figure*}

\section{Introduction}

Automatic generation of chest X-ray reports from medical images represents a significant challenge in multi-modal artificial intelligence \citep{iuxray, chexpert, mimiccxr, rexrank}. Effective AI systems in this domain can substantially reduce labor costs and help standardize radiological interpretations. Recent multi-modal large language models (MLLMs), including ChexAgent \citep{chexagent} and MAIRA-2 \citep{maira2}, have achieved competitive performance on natural language generation metrics. However, human evaluations indicate these models often hallucinate details or omit critical clinical information \citep{zhang2024generalist, tu2024towards, wu2024hallucinationbenchmarkmedicalvisual}.

To overcome these limitations and improve clinical accuracy, reinforcement learning (RL) presents a promising alternative to supervised fine-tuning (SFT). Unlike SFT, which is sensitive to specific phrasing, RL directly optimizes clinical efficacy metrics, allowing greater flexibility in expression and prioritizing essential clinical content. Previous studies \citep{zhou2024large, yang2025aligning} have explored Proximal Policy Optimization (PPO) for report generation, but in LLM settings PPO is costly to optimize: it requires an additional value function comparable in size to the policy, and token-level value estimation is difficult under end-of-sequence rewards. To address this, Group Relative Policy Optimization (GRPO) \citep{shao2024deepseekmath} removes the value function by computing advantages based on normalized rewards relative to peer samples.
However, GRPO, as an on-policy RL method, inherently depends on the model's current capabilities, limiting effective skill expansion. Our experiments using QWen2.5-VL-3B \citep{bai2025qwen2} reveal that this constraint leads to low-quality outputs, frequently defaulting to generic "no finding" statements. Consequently, advantage estimates remain flat, and training performance quickly plateaus, as demonstrated in Fig.~\ref{fig:val_trend}.

To address the exploration dilemma, we propose EditGRPO, a variant of GRPO that edits generated report candidates during rollout by injecting clinically correct information from reference reports. Edits include deleting or replacing incorrect (false-positive) findings and adding missing (false-negative) findings as shown in Fig. \ref{fig:overview}. These edits
keep the model close to its existing policy while enabling off-policy guidance
to bring its generations closer to the reference report.

Our work introduces three major contributions:
\begin{itemize}
    \item We propose EditGRPO, a mixed-policy RL algorithm that balances exploration and imitation, effectively overcoming the sampling efficiency bottleneck in report generation. 
    \item We explore a range of editing strategies and introduce a similarity-based, sentence-level editing approach that mitigates large policy shifts by making minimal yet reward-improving modifications.
    \item We implement EditGRPO along with a two-stage training strategy across four multi-view and longitudinal chest X-ray report datasets, outperforming three widely used medical MLLMs. Furthermore, out-of-domain evaluation reveals that EditGRPO generalizes well to unseen data distributions.
\end{itemize}




\section{Motivation for Post-Rollout Edits}

A key obstacle to applying reinforcement learning to chest X‑ray report generation is a shortcut bias: because normal studies vastly outnumber abnormal ones, a policy can maximize its expected reward by emitting a generic ``No abnormality'' statement. 
This imbalance drives policy collapse, with the model converging to a single high‑probability ``normal'' mode—an effect that becomes even more pronounced in low‑resource settings (see Table~\ref{tab:main_results_mimic_cxr}, where training with pure GRPO results in low CheXbert Macro-F1 scores.). Unlike classification tasks, a radiology report is a holistic narrative that may describe an arbitrary subset and count of findings, so straightforward class‑frequency re‑weighting \citep{lin2017focal} may not be practical for all findings simultaneously. To counter this collapse we introduce \textit{post‑rollout edits}: after sampling a candidate report $o_i$ from the policy, we inject expert corrections based on the gold-standard report, recompute the edited trajectory’s log‑probability, and use the resulting adjusted advantage to steer the policy gradient toward clinically faithful outputs. Concretely, within each prompt's group we interleave unedited rollouts with minimally edited counterparts. This creates intrinsic raw-versus-edited contrasts that enlarge group-relative reward gaps on clinically specific findings, yielding a sharper advantage signal while staying close to the current policy. To prevent policy drift, we apply this editing at the sentence level,, which also help the model absorb fine‑grained corrections throughout the long‑context report.

\section{Method}

The formulation of GRPO is reviewed in Equation \ref{eq:grpo} in the Appendix \ref{appx:grpo}.  Within a group
of responses $o_i$, $i = 1, \ldots, G$ to a prompt $q$,
GRPO generally uses a normalized advantage at every token $t$:

{\small
\begin{equation}
\hat{A}_{i,t} = 
\frac{
R(q, o_i) - \text{mean}\left(\left\{ R(q, o_1), \ldots, R(q, o_G) \right\}\right)
}{
\textcolor{red}{
\text{std}\left(\left\{ R(q, o_1), \ldots, R(q, o_G) \right\}\right)
}}
\label{eq:advantage}
\end{equation}
}

\noindent but Dr. GRPO \citep{drgrpo} revealed that vanilla GRPO introduces optimization biases. For example, longer responses are penalized less due to their larger $|o_i|$, leading the policy to favor lengthier but potentially incorrect outputs. Additionally, a question-difficulty bias arises from the standard deviation term in the advantage function, where questions with low standard deviation (i.e., very easy or very hard, where rewards are mostly 1 or 0) are disproportionately weighted during policy updates. To mitigate these issues, EditGRPO adopts their proposed modification and \textcolor{red}{removes this bias-inducing quotient} as in Dr. GRPO, to use {\em unnormalized advantage.} By default, EditGRPO samples $\frac{G}{2}$ responses \( o_i \), and fills the other half of the batch with edited versions $o^\prime_i$ of these.

\smallskip

\noindent\textbf{Editing Rule.}  
Given a generated report \(\mathbf{x}\)
and a reference report \(\mathbf{y}\),
the RaTE-NER model \citep{zhao-etal-2024-ratescore} extracts entities from each, together with vector embeddings and labels indicating presence or absence.  Fix a cosine similarity
threshold $\tau$ (see Table \ref{tab:ablation_tau} for the hyperparameter sensitivity analysis).
For each reference sentence $y_j$, let $E[y_j]$  to be the collection of all entities $e$ with $\cos(e,e') > \tau$ for some $e^\prime \in y_j$.  An entity $e \in x$ is {\em spurious} if it is not in any $E[y_j]$ with a matching presence label. RadGraph can also identify edit candidates but is sensitive to phrasing (see Table \ref{tab:ablation_editing} in the appendix).

\begin{enumerate}[label=(\alph*), leftmargin=12pt]
  \item \textbf{Mislabeled Entities.}  
    If there exists an entity \(e\) in \(x_i\) and \(y_j\) with conflicting presence labels,  
    select one such conflict uniformly at random and replace \(x_i\) with \(y_j\).

  \item \textbf{False‑positive replacement.}  
    Otherwise, if some sentence \(x_i\) contains a set of spurious entities 
    \(E_{\mathrm{fp}}\not\subseteq E[y_j]\) for all \(j\) at threshold $\tau$,  
    compute
    \[
      s^* \;=\; \arg\max_{j:\,y_j\neq x_i}
      \frac{1}{|E_{\mathrm{fp}}|}\sum_{e\in E_{\mathrm{fp}}}\max_{e'\in E[y_j]}\cos(e,e').
    \]
    If 
    \(\frac{1}{|E_{\mathrm{fp}}|}\sum_{e\in E_{\mathrm{fp}}}\max_{e'\in E[y_{s^*}]}\cos(e,e')\ge \tau\),
    replace \(x_i\) with \(y_{s^*}\).

  \item \textbf{False‑positive deletion.} Otherwise, if no replacement was made and there is  still a sentence \(x_i\) with false‑positive entities, delete it.

  \item \textbf{False‑negative augmentation.}  
    Otherwise, if there is an entity in \(\mathbf{y}\) missing from \(\mathbf{x}\),  
    select one reference sentence \(y_j\) containing it and \emph{append} \(y_j\) to the end of \(\mathbf{x}\).

  \item \textbf{Termination.}  
   If none of the above applies, return \(\mathbf{x}\); if it is empty, replace it with any unused false‑negative sentence \(y_j\). 

\end{enumerate}


\begin{table*}[ht]
\centering
\resizebox{\textwidth}{!}{%
\begin{tabular}{l|c*{7}{c}}
\toprule
\textbf{Method}
  & \textbf{Micro‐F1‑14}
  & \textbf{Macro‐F1‑14}
  & \textbf{Micro‐F1‑5}
  & \textbf{Macro‐F1‑5}
  & \textbf{RadGraph F1}
  & \textbf{RaTE}
  & \textbf{GREEN}
  & \textbf{Avg.} \\
\midrule
\rowcolor{gray!20} 
\multicolumn{9}{c}{MIMIC-CXR} \\
\midrule
MAIRA-2 (7B) & 0.5154 &0.3557 &0.5531 &0.4695 & 0.2104 & 0.3042 &\textbf{0.5042} &0.4161 \\
ChexAgent (8B) &0.2981 &0.1772 &0.3593 &0.2363 &0.1544 &0.4370 &0.2359 &0.2712 \\
MedGemma (4B) &0.4742 &0.3462 &0.5222 &0.4752 & 0.1113 &0.4680  & 0.2253 &0.3746  \\
\midrule
{\em Qwen2.5-VL-3B} & & & & & & & & \\
\hspace{0.5em}SFT (\textit{ep3})                           & 0.5144 & 0.3344 & 0.5625 & 0.4737 & 0.2937 & 0.5434 & 0.3583 & 0.4401 \\
\hspace{0.5em}GRPO                          & 0.1381 & 0.0966 & 0.1167 & 0.0981 & 0.0905 & 0.4067 & 0.2676 & 0.1735 \\
\hspace{0.5em}SFT(\textit{ep2}) + GRPO                    & 0.4781 & 0.3042 & 0.5376 & 0.4527 & \textbf{0.2963} & 0.5411 & 0.3497 & 0.4228   \\
\hspace{0.5em}SFT(\textit{ep2}) + Dr.\ GRPO               & 0.5470 & 0.3481 & 0.6050 & 0.5023 & 0.2959 & 0.5460 &    0.3627    & 0.4528 \\
\hspace{0.5em}SFT(\textit{ep2}) + EditGRPO (\textit{para})& 0.5353 & 0.3458 & 0.5946 & 0.4986 & 0.2546 & 0.5257 & 0.3114        &0.4380   \\
\hspace{0.5em}SFT(\textit{ep2}) + EditGRPO (\textit{norm}) & 0.5467  & \textbf{0.3789}  & 0.5986 
& \textbf{0.5388} & 0.2721 & 0.5441 & 0.3500   & 0.4613   \\
\hspace{0.5em}SFT(\textit{ep2}) + EditGRPO                   & \textbf{0.5594} & 0.3674 & \textbf{0.6124} & 0.5242 & 0.2841 & \textbf{0.5712} &        0.3825 & \textbf{0.4716}  \\
\midrule
\rowcolor{cyan!20} 
\multicolumn{9}{c}{RexGradient} \\
\midrule
{\em Qwen2.5-VL-3B} & & & & & & & & \\
\hspace{0.5em}SFT (\textit{ep2})                          & 0.4035 & 0.1969 & 0.2273 & 0.1748 & 0.3294 & 0.5811 &0.4430  & 0.3366 \\
\hspace{0.5em}SFT (\textit{ep1}) + EditGRPO                  &\textbf{0.4061}  &\textbf{0.2159}   &\textbf{0.2867}   &0.2168   &0.2986    &\textbf{0.5981}  &  \textbf{0.4756}  & \textbf{0.3568}  \\
\bottomrule
\end{tabular}%
}
\caption{Performance of various training variants and medical \textit{SoTA} models across clinical metrics on two large-scale RRG datasets: MIMIC-CXR and RexGradient. \textbf{\textit{Note:}} EditGRPO is built on the Dr.\ GRPO and utilizes sentence‑level edits by default; \textit{norm} indicates normalized advantage are utilized; \textit{para} means edits are performed on the paragraph (or report) level. \textit{ep*} denotes the training epoch index.}
\label{tab:main_results_mimic_cxr}
\end{table*}

\section{Experiments}

\subsection{Experimental Setting}
\noindent \textbf{Data.} We focus on MIMIC-CXR \citep{mimiccxr}, a well-studied large-scale dataset, for both training and evaluation. Additionally, we incorporate the newly released large-scale RexGradient \citep{zhang2025rexgradient} dataset to further validate the effectiveness of EditGRPO. We also study two smaller benchmark datasets, IU-XRay \citep{iuxray} and PadChest-GR \citep{de2025padchest}, both for training and for out-of-domain generalization. 
See Appendices \ref{appx:data} and \ref{appx:ablation} for more details.

\smallskip
\noindent \textbf{Models.} Qwen-VL-2.5 (3B) \citep{bai2025qwen2} is used for training report generation models. For a comparative baseline on MIMIC-CXR, we include results from other models supporting \emph{multi-view and longitudinal} inputs, such as ChexAgent \citep{chexagent}, MAIRA-2 \citep{maira2}, and MedGemma \citep{sellergren2025medgemma}. The performance of the baselines reported in this paper is obtained through our own inference using the curated \emph{multi-view and longitudinal} datasets. The model's performance may be influenced by the prompt.

\smallskip
\noindent \textbf{Training.} Because of the
time and resources required for reinforcement learning, we compare results
for MIMIC-CXR at step 1000, corresponding to about 1/4 of the training data,
even though performance continues to improve as shown in Fig. \ref{fig:val_trend}.  For RexGradient, we use the checkpoint at step 600, representing roughly 15\% of the data scale.
For the smaller datasets,
we train for one full epoch. 
We adopt a composite reward strategy, as different clinical efficacy metrics evaluate report quality from complementary perspectives. 

As primary reward signals, we adopt the following metrics and combine them as an unweighted sum, using Micro-F1 rather than Macro-F1 due to the sparsity and weak supervision of macro-level rewards across most datasets (see Appendices \ref{sec:reward} and \ref{appx:training}). The resulting composite reward is:

{\small
$$R = \mathrm{RadGraph\text{-}F1} + \mathrm{CheXbert\text{-}Micro\text{-}F1\text{-}14} + \mathrm{RaTE}.$$}


\smallskip
\noindent \textbf{Evaluation.} We evaluate the generated reports using radiology-specific metrics, following established protocols. These include RadGraph-F1 \citep{radgraph}, CheXbert scores (Micro/Macro-F1-14/5) \citep{smit-etal-2020-combining}, RaTE Score \citep{zhao-etal-2024-ratescore}, and GREEN \citep{ostmeier2024green}. These clinical metrics are designed to emphasize the accuracy of medical findings, with a focus on detecting and assessing clinically relevant entities. We applied a two-sided Wilcoxon signed-rank test, which confirmed statistically significant gains for EditGRPO on most metrics (Table~\ref{tab:delta_consolidated}).

\begin{figure*}[!htb]
    \centering
    \includegraphics[width=\linewidth]{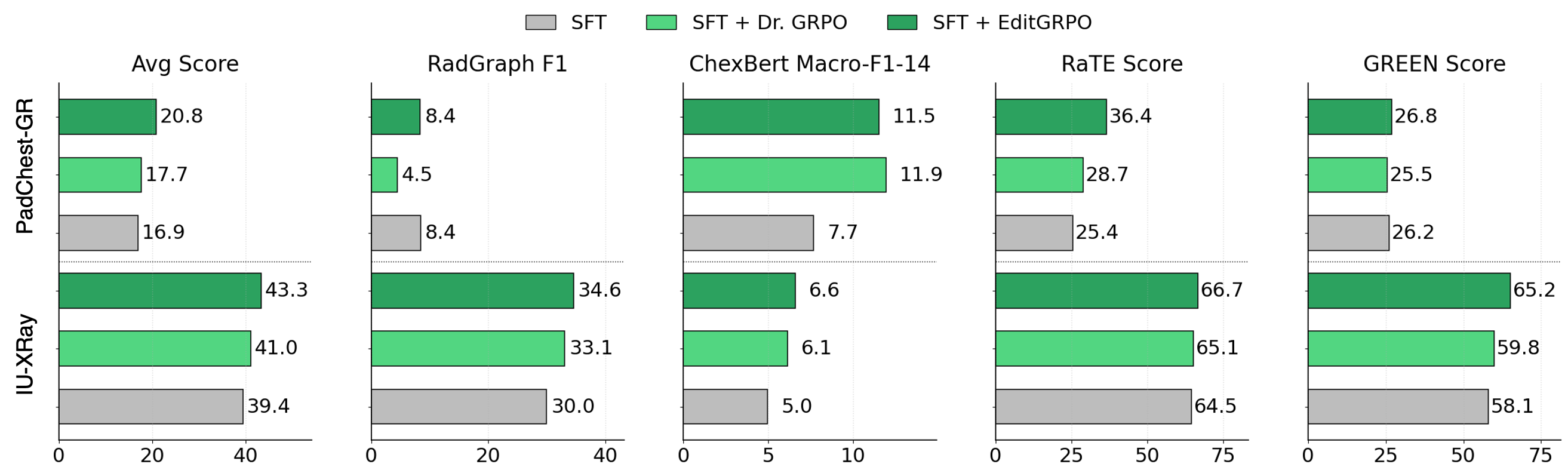}
    \caption{Performance (\%) of different training strategies on two small-scale datasets: IU-XRay and PadChest-GR.}
    \label{fig:main_results_others}
\end{figure*}

\subsection{Main Results} 

\smallskip
\noindent \textbf{SFT is necessary before RL for chest X-ray report generation.} Table \ref{tab:main_results_mimic_cxr} presents detailed results for the MIMIC-CXR dataset, alongside the performance of SFT and EditGRPO methods on RexGradient. GRPO alone achieves an average score of only 17.35\%, demonstrating limited effectiveness without prior SFT training. However, incorporating an initial SFT phase significantly boosts performance, yielding approximately a 25\% improvement. The same conclusion is also supported by related works \citep{guo2025deepseek, fan2025chestx, liu2025x}.

\smallskip
\noindent \textbf{EditGRPO surpasses SFT and other RL variants.} The SFT+EditGRPO model (or its normalized advantage variant) lead all
metrics except RadGraph F1, which is sensitive to the exact wordings
prioritized by SFT training.  Accounting for semantic equivalence, we see
a 5.0\% improvement in CheXbert Micro-F1-5, 4.5\% improvement in
CheXbert Micro-F1-14, and 2.8\% improvement in RaTEScore with SFT+EditGRPO.
For the macro variants of the CheXbert metrics, the normalized advantage
variant of EditGRPO performs slightly better, although its overall performance is generally lower. 
An ablation where EditGRPO modifies the entire paragraphs rather than the individual sentence (limited to one batch element per query) significantly reduces performance. This highlights the importance of small, localized edits closely aligned with the current policy.

\smallskip
\noindent \textbf{EditGRPO enables general-domain models to detect abnormalities more accurately than existing domain-specific models.} As shown in Table~\ref{tab:main_results_mimic_cxr}, applying EditGRPO to Qwen-VL-2.5 (3B) achieves state-of-the-art performance across all metrics except the GREEN score. Notably, we observe substantial gains in CheXbert-F1 scores, which are directly associated with the model's ability to detect common abnormalities. Specifically, SFT + EditGRPO yields improvements over MAIRA-2 of 4.4\%, 1.2\%, 5.9\%, and 5.5\% across four CheXbert-F1 variants, respectively. 

\smallskip
\noindent \textbf{SFT-then-EditGRPO performs effectively on small-scale datasets.} Fig.~\ref{fig:main_results_others} demonstrates the effectiveness of EditGRPO on two smaller datasets: IU-XRay and PadChest-GR. We primarily focus on RadGraph-F1, RaTE Score, CheXbert-Macro-F1-14, and GREEN Score for evaluation. On IU-XRay, the SFT+EditGRPO model achieves improvements over SFT by 4.6\% (RadGraph-F1), 2.2\% (RaTE Score), 1.6\% (CheXbert-Macro-F1-14), and 7.1\% (GREEN Score). Similarly, on PadChest-GR, SFT+EditGRPO outperforms SFT by 3.8\% (ChexBert-Macro-F1-14), 11.0\% (RaTE Score), and 0.6\% (GREEN Score), although the improvement in CheXbert-Macro-F1-14 is marginal. Comprehensive metrics are detailed in Table \ref{tab:additional_results_small_scale}. 

\smallskip
\noindent \textbf{RL demonstrates better generalization.} Table \ref{tab:main_generalization} compares the average cross‐metric performance (Avg. Score) of models trained on MIMIC‑CXR and then evaluated on two external datasets, IU‑XRay and PadChest‑GR, using three training strategies. On PadChest‑GR, the SFT baseline achieves an Avg. Score of 0.174, which rises to 0.193 (+11\%) when Dr.GRPO is applied and further to 0.260 (+49\%) with EditGRPO. A similar but less pronounced trend appears on IU‑XRay: SFT starts at 0.403, Dr. GRPO improves it to 0.416 (+3\%), and EditGRPO brings it up to 0.435 (+8\%). 
Moreover, EditGRPO consistently yields the highest Avg. Score across both target domains, demonstrating its superior robustness in out‐domain generalization (see Table \ref{tab:additional_results_generalization} for the full breakdown of evaluation metrics).

\begin{table}[]
    \centering
    \resizebox{0.49\textwidth}{!}{
    \begin{tabular}{l|ccc}
        \toprule
         \textbf{Dataset} & \textbf{SFT} & \textbf{SFT + Dr. GRPO} & \textbf{SFT + EditGRPO} \\
         \midrule
         IU-XRay &0.4026   &0.4160   &\textbf{0.4348}   \\
         PadChest &0.1742   &0.1926  &\textbf{0.2595}   \\
         \bottomrule
    \end{tabular}
    }
    \caption{Average score across all evaluation metrics for models trained on MIMIC-CXR.}
    \label{tab:main_generalization}
\end{table}







\section{Conclusion}
Our EditGRPO framework makes it possible to optimize diverse clinical efficacy metrics in a GRPO-based framework, addressing the failure of pure GRPO to adequately explore.  Even with a simple MLLM architecture pretrained on general-domain data, our technique achieves gains in multiple rewards with either large or small training sets.  Our technique is effective even in challenging multi-view, longitudinal image settings.  Furthermore, SFT+EditGRPO models from a large training set generalize better than SFT or SFT+Dr.GRPO to small, out-of-domain data.  Although our focus is on the training algorithm, we expect advances in architectures or training data could be integrated to further improve performance.


\section{Limitation}
Although our proposed EditGRPO method demonstrates the robust potential of reinforcement learning to improve the clinical efficacy of chest X-ray report generation, several limitations remain. First, we did not extend training to larger models (e.g., 7B or 32B parameters), primarily due to the multiplicative increase in computational and time requirements. This constraint is especially relevant in our settings involving multi-view and longitudinal data, where multiple image inputs are required and resource efficiency becomes critical. Second, while the current clinical metrics used in evaluation are generally reliable, they may not fully align with expert radiologist assessments. Additional human evaluation will be necessary to more comprehensively validate the quality and safety of reports generated by models trained with EditGRPO. 
Thirdly, our rule- and NER-based editing prioritizes precision and auditability but may be brittle to paraphrase, clinical severity nuances, and discourse placement. While stronger LLM editors could improve fluency and insertion, they raise hallucination/negation risks and compute cost; a constrained, entity/negation-aware LLM editor is a promising direction for future work.
Lastly, this work focuses specifically on chest X-ray report generation. The applicability of EditGRPO to other medical imaging modalities, such as CT or MRI, remains an open question and is a promising direction for future research.

\bibliography{grpo2,grpo}

\appendix

\section{Related Work}

\subsection{Radiology Report Generation}

Recent research has focused on transformer-based architectures for medical imaging analysis \citep{llavamed}.
Some of these systems are differentiated by their ability to support
multi-image inputs, including MAIRA-2 \citep{maira2} and ChexAgent \citep{chexagent}.  The recent leaderboard RexRank \citep{rexrank}
studies only a single-image setting, and the latest leaders such as MedVersa \citep{medversa} focus on integrating multiple modules for detection and
segmentation and a large-scale data wrangling effort.
In contrast, we focus on training algorithm improvements for a commonly used MLLM
architecture, rather than an architecture or data contribution.

\subsection{Reinforcement Learning}
To further enhance clinical accuracy, some methods incorporate RL to optimize for task-specific rewards, such as capturing ``clinically relevant'' features \citep{liu2019clinically, kaur2022cadxreport, zhou2024large} or maintaining logical consistency \citep{delbrouck2022improving, miura2021improving}. Rule-based reinforcement learning techniques, exemplified by Group Relative Policy Optimization (GRPO), have shown strong potential for large-scale RL applications, particularly in tasks such as mathematical reasoning and code generation, and have recently been extended to multimodal setting \citep{grpomultimodal}. In the medical domain, GRPO has been explored in visual question answering tasks \citep{pan2025medvlm, fan2025chestx, lai2025med}. However, to the best of our knowledge, it has not yet been applied to radiology report generation. 

\subsection{Clinical Evaluation Metrics}
Evaluating the quality of generated radiology reports is non-trivial. Early works adopted general-domain natural language processing metrics such as ROUGE \citep{lin-2004-rouge}, BLEU \citep{papineni-etal-2002-bleu}, and BERTScore \citep{bertscore}. While these metrics are widely used for text evaluation, they treat differences in wording the same as clinically significant errors, failing to reflect medical accuracy. To address this limitation, clinically informed evaluation metrics, such as CheXbert \citep{smit-etal-2020-combining}, RadGraph \citep{radgraph}, GREEN \citep{ostmeier2024green}, and RaTEScore \citep{zhao-etal-2024-ratescore}, have been proposed to better assess clinical correctness and utility. 
CheXbert is based on multi-label classification results for 5 or 13 diseases (along with one extra ``normal'' label).
RadGraph considers literal entity agreement considering the positive or negative
context of each entity.  GREEN judges recall and precision errors by LLM
prompting.  RaTEScore is inspired by RadGraph but less sensitive to phrasing
by an F1-like computation which allows semantic matching between entities based on a cosine similarity.

\section{Group Relative Policy Optimization} \label{appx:grpo}

Let \( P(Q) \) denote the distribution over questions (images and prompts) used for training, where \( q \) is a sampled question in the current iteration. Let \( \pi_{\theta_{\text{old}}} \) and \( \pi_{\theta_{\text{new}}} \) denote the old policy and current (new) policy, respectively, where \( o \) is a complete response sampled from a policy. Let \( \pi_{\theta_{\text{ref}}} \) denote the reference policy, which in practice is the frozen base MLLM. Let \( G \) be the number of responses sampled per question in each iteration. The GRPO objective is given by:

\scalebox{0.8}{%
  \parbox{\linewidth}{%
    \begin{align}
    \mathcal{J}_{\text{GRPO}}(\theta) 
	    = \, & \mathbb{E}_{q \sim P(Q), \{o_i\}_{i=1}^{G} \sim \pi_{\theta_{\text{old}}}(O|q)} \Bigg[ \frac{1}{G} \sum_{i=1}^{G} {\frac{1}{|o_i|}} \sum_{t=1}^{|o_i|} \Bigg( \nonumber \\
	    & \min \Bigg( \frac{\pi_{\theta_{\text{new}}}(o_{i,t} \mid q)}{\pi_{\theta_{\text{old}}}(o_{i,t} \mid q)} A_{i,t}, \nonumber \\
	    & \quad \text{clip} \left( \frac{\pi_{\theta_{\text{new}}}(o_{i,t} \mid q)}{\pi_{\theta_{\text{old}}}(o_{i,t} \mid q)}, 1 - \epsilon, 1 + \epsilon \right) A_{i,t} \Bigg) \Bigg) \nonumber \\
    & - \beta \, \mathbb{D}_{\text{KL}}\left( \pi_{\theta_{\text{new}}} \,\|\, \pi_{\theta_{\text{ref}}} \right) \Bigg],
    \label{eq:grpo}
    \end{align}
  }%
}
, where \( \frac{\pi_{\theta}(o_i \mid q)}{\pi_{\theta_{\text{old}}}(o_i \mid q)} \) is the policy ratio, and \( A_i \) is the estimated advantage defined in Equation \ref{eq:advantage}, and \( \epsilon \) is the clipping threshold for policy updates.

\section{Reward Design}
\label{sec:reward}

\paragraph{Reasoning or no-thinking?} Most existing MLLM works that utilize the GRPO algorithm aim to incentivize reasoning capabilities in the VQA setting \citep{lai2025med, pan2025medvlm, liu2025visual, huang2025vision, fan2025chestx, liu2025x}. To achieve this, they incorporate a format score that encourages the model to enclose its reasoning process within \textit{<think>} and \textit{</think>} tags. However, for report generation, explicit self‑rationalization is unnecessary: the radiology report itself inherently embodies the trackable reasoning of imaging analysis. It performs semantic mapping from image features to clinical description and lays out the inferred relationships among findings and their differential diagnoses. Therefore, we do not incorporate a format reward and instead conduct ``no-thinking'' GRPO setup.



\noindent \paragraph{Clinical efficacy rewards.}

While other metrics, such as GREEN, show stronger alignment with human evaluations \citep{ostmeier2024green}, they rely on LLM-based evaluations, which incur significantly higher inference latency and resource costs.
The metrics are computed using their official and standardized implementations: 
\textsc{RadGraph-F1}\footnote{\url{https://pypi.org/project/radgraph/0.1.2/}}, 
\textsc{CheXbert-F1}\footnote{\url{https://pypi.org/project/f1chexbert/}}, 
\textsc{RaTE Score}\footnote{\url{https://pypi.org/project/RaTEScore/0.5.0/}}, and 
\textsc{GREEN}\footnote{\url{https://pypi.org/project/green-score/0.0.8/}}.

\begin{table*}[ht]
\small
\centering
\begin{tabular}{l|cc|cc|cc}
\toprule
\multirow{2}{*}{\textbf{Dataset}} & \multicolumn{2}{c|}{\textbf{\# Samples}} & \multicolumn{2}{c|}{\textbf{\# Images}} & \multicolumn{2}{c}{\textbf{\% Has Prior}} \\
\cmidrule(lr){2-3} \cmidrule(lr){4-5} \cmidrule(lr){6-7}
 & Train & Test & Train & Test & Train & Test \\
\midrule
MIMIC-CXR    & 146{,}893 & 2{,}231 & 4.17 ± 0.63 & 4.30 ± 0.62 & 42.59 & 68.60 \\
RexGradient  & 139{,}884 & 9{,}992 & 1.69 ± 0.63 & 1.69 ± 0.63 & 0 & 0 \\
IU-Xray      & 2{,}365   & 590     & 2.00 ± 0.00 & 2.00 ± 0.00 & 0 & 0 \\
PadChest-GR  & 3{,}640   & 915     & 1.32 ± 0.47 & 1.31 ± 0.46 & 32.40 & 31.91 \\
\bottomrule
\end{tabular}
\caption{\label{tab:data_stat}The datasets used for training and evaluating EditGRPO include statistics such as the proportion of samples containing prior studies and the total number of images. For RexGradient, cases containing more than six images were excluded due to resource constraints. Note that The numbers reported in this table reflect the dataset statistics after preprocessing, rather than the raw data.}
\end{table*}

\section{Training Configurations} \label{appx:training}


Training begins with a general-domain MLLM with trainable parameters $\theta$, following a two-stage approach: SFT followed by RL. This strategy allows the model to first adapt to the chest X-ray domain and subsequently perform effective sampling during the RL phase.

\paragraph{Stage 1: SFT for domain adaptation.} We train the model using cross-entropy loss until both the loss and clinical metrics stabilize. Formally, for each \textit{multi-view and longitudinal} chest X-ray image set $\mathbf{X}_v$, we append a text prompt $\mathbf{X}_q$ containing clinical context information such as indication, technique, and comparison. The MLLM is then trained to predict the target report tokens using its original autoregressive objective. Specifically, given a target findings section $\mathbf{X}_a$ of length $L$, the model is optimized to maximize the likelihood:

{\small
$$\mathcal{L}_{\mathrm{SFT}}(\theta) = -\sum_{i=1}^L
\log p_{\theta}\bigl(x_i \mid
\mathbf{X}_v,\,
\mathbf{X}_{q,<i},\,
\mathbf{X}_{a,<i}
\bigr).$$}

\paragraph{Stage 2: RL for clinical improvement.} To prevent model collapse, we use the checkpoint from the epoch prior to convergence as the starting point for RL training. The optimization objective is described in Equation \ref{eq:grpo}.

\paragraph{Prompts.} An prompt example with two images are shown in the following: 

\lstset{
  basicstyle=\footnotesize\ttfamily,   
  breaklines=true,                     
  breakatwhitespace=true,              
  columns=fullflexible,                
  keepspaces=true,                     
  frame=single,                        
  xleftmargin=1em,                     
  xrightmargin=1em,                    
  backgroundcolor=\color{gray!5},      
  rulecolor=\color{gray!50},           
  upquote=true                         
}

\begin{lstlisting}[linewidth=\columnwidth]
<image><image>

As a radiologist assistant, your task is to interpret a chest X-ray study. Given the current <view> image, and the prior <view> image.

Please provide a detailed description of the findings from the image(s) in the current study. If there are prior studies available, please incorporate relevant details from those as well in your analysis.

INDICATION: <INDICATION>
TECHNIQUE:  <TECHNIQUE>
COMPARISON:  <COMPARISON>
\end{lstlisting}

\paragraph{Hyperparameters.} In the reinforcement learning setup, we use a rollout batch size of 32. Image inputs are constrained to a maximum of 401,408 pixels and a minimum of 262,144 pixels. We employ the AdamW \citep{loshchilov2017decoupled} optimizer with a learning rate of 1.0 e-6 and a weight decay of 1.0 e-2. During the rollout phase, the generation temperature is set to 1.0, and we sample $\frac{G}{2}=5$ responses per prompt. To reduce memory usage, we enable gradient checkpointing \citep{chen2016training} and apply Fully Sharded Data Parallelism (FSDP) \citep{zhao2023pytorch}. Additionally, vLLM \citep{kwon2023efficient} is used to accelerate inference and response generation. 

\paragraph{Computation.} Training Qwen2.5-VL-3B is conducted on a computational infrastructure equipped with NVIDIA A100 GPUs (80GB memory). For \textsc{MIMIC-CXR}, \textsc{EditGRPO} training up to step 1000 is estimated to take approximately 5 days using four GPUs. For \textsc{RexGradient}, training is estimated at 3 days with four GPUs. In contrast, training on \textsc{IU-Xray} requires roughly 5 hours using two GPUs, while \textsc{PadChest-GR} takes approximately 8 hours using two GPUs.

\section{Datasets} \label{appx:data}

For all datasets, we follow the official train-test splits. For MIMIC-CXR and PadChest-GR, we construct multi-view and longitudinal inputs based on metadata. Specifically, we retain both frontal and lateral views and identify prior exams using chronological order. Prior reports are excluded from the input to minimize token costs—given the resource-intensive nature of multi-image settings—and to prevent shortcut learning, such as copying content from earlier reports.

It is worth noting that MIMIC-CXR includes a ``Comparison'' section but does not provide exact identifiers for prior exams, making precise longitudinal matching infeasible. Therefore, we use the most recent prior exam to build sequential image inputs. For IU-XRay and RexGradient, we only curate multi-view inputs due to limitations in longitudinal metadata. 

The detailed data statistics is shown in Table \ref{tab:data_stat}.

\paragraph{Data license and use agreement.} Our study employs four fully de‑identified, publicly released chest radiograph collections—MIMIC‑CXR \citep{mimiccxr, johnson2019mimic}, PadChest‑GR \citep{de2025padchest} (English version), IU‑Xray \citep{iuxray}, and RexGradient \citep{zhang2025rexgradient}—each governed by permissive open‐use terms that together guarantee ethical compliance and reproducibility. The MIMIC‑CXR dataset is accessed via PhysioNet under its data use agreement and CITI ``Data or Specimens Only Research'' certification \cite{goldberger2000physiobank}. PadChest‑GR is distributed under the PADCHEST Dataset Research Use Agreement\footnote{\url{https://bimcv.cipf.es/bimcv-projects/padchest/padchest-dataset-research-use-agreement/}}, which grants free access for academic research only and explicitly prohibits redistribution or commercial use. IU‑Xray is released under a CC BY‑NC‑ND 4.0 license, which permits non‑commercial redistribution with attribution but prohibits both commercial reuse and the creation of derivative works. RexGradient is available via Hugging Face\footnote{\url{https://huggingface.co/datasets/rajpurkarlab/ReXGradient-160K}}, and is released under a Non‑Commercial Data Access and Use Agreement that restricts use to non‑clinical, non‑commercial research while requiring proper attribution to Harvard Medical School and Gradient Health. No patient can be re‑identified from any of these sources, and all secondary analyses conform strictly to each dataset’s specific attribution, share‑alike, and noncommercial clauses. By restricting our work to these four open‑license datasets, we uphold the highest standards of data privacy, transparency, and legal clarity in ethical AI research.



\section{Additional Analysis} \label{appx:ablation}

\begin{figure}
    \centering
    \includegraphics[width=\linewidth]{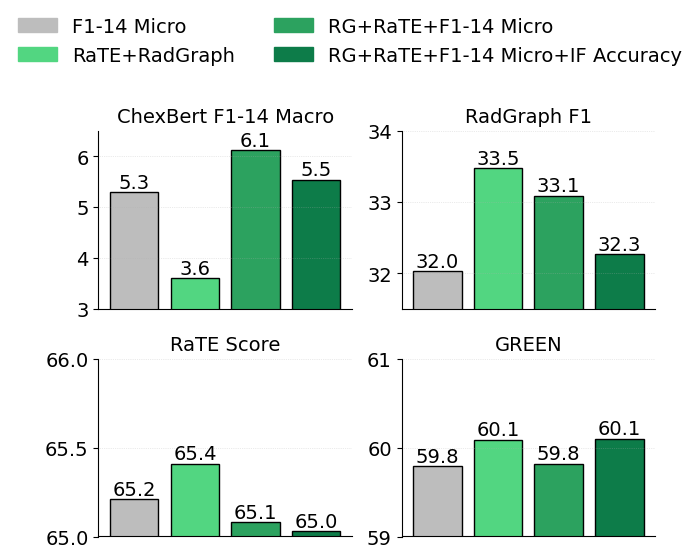}
    \caption{Influence of reward design on the IU-XRay dataset under the SFT + Dr.GRPO setting. \textit{RG} denotes the RadGraph reward, and \textit{IF} denotes the inverse‑frequency reward, which assigns higher scores when a rare condition is hit according to the label distribution across the 14 CheXpert classes \citep{chexpert} of the training data.}
    \label{fig:reward_design_iu_xray}
\end{figure}

\begin{figure}
    \centering
    \includegraphics[width=0.9\linewidth]{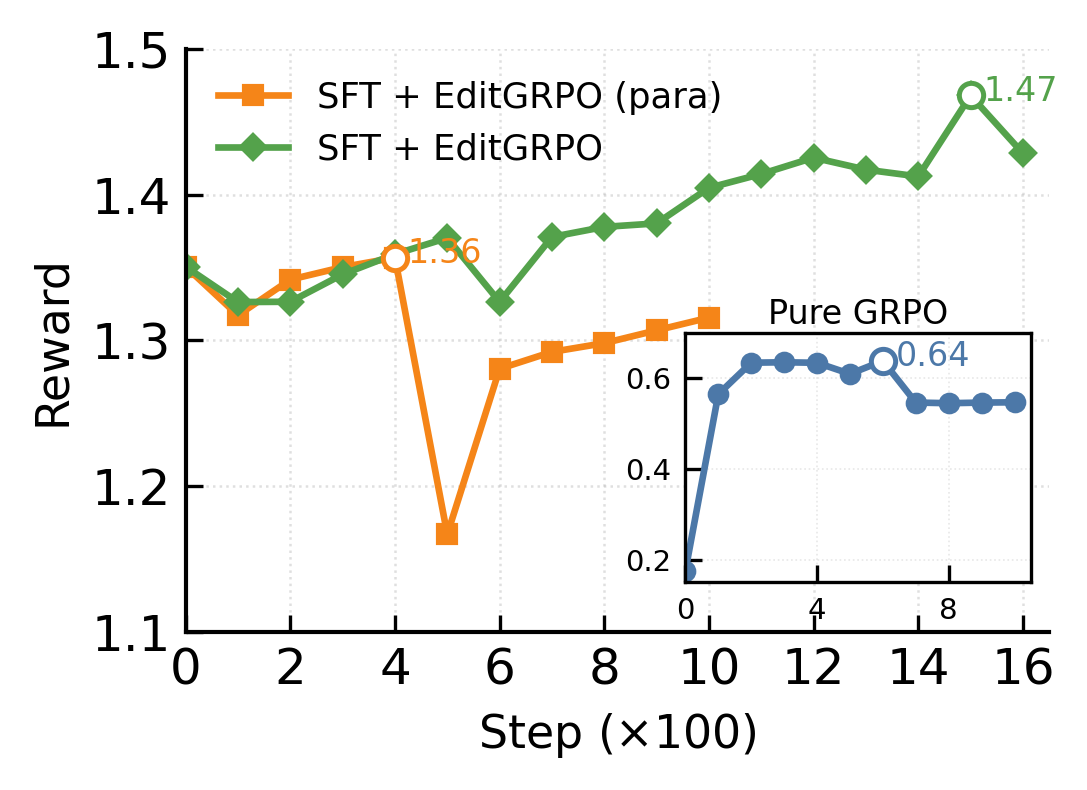}
    \caption{Reward gains (RadGraph + RaTE + Chexbert-Micro-F1-14, the maximum is 3) over training step on MIMIC-CXR.}
    \label{fig:val_trend}
\end{figure}

In this section, we present comprehensive evaluation metrics on IU-XRay and PadChest-GR, as shown in Table~\ref{tab:additional_results_small_scale}. In addition, we conduct ablation studies on both reward design (see Fig. \ref{fig:reward_design_iu_xray}) and post-rollout editing strategies (see Fig. \ref{fig:val_trend}) to analyze their individual contributions to overall performance. Detailed results of the statistical significance tests are presented in Table~\ref{tab:delta_consolidated}.

\paragraph{Composite reward is effective.}
Figure~\ref{fig:reward_design_iu_xray} compares four reward schemes on IU‑XRay. When optimizing solely for RaTE and RadGraph, the ChexBert-Macro-F1‑14 score falls sharply from 5.3\% to 3.6\%, indicating that these metrics alone do not drive macro‑F1 improvements. Adding the F1‑14 Micro component not only recovers but surpasses the baseline, boosting macro‑F1 to 6.1; the subsequent inclusion of IF Accuracy yields a slight decrease to 5.5, suggesting diminishing returns from additional reward terms.  The decreases in RadGraph-F1, RaTE Score, and GREEN by adding the Chexbert-F1-14 rewards to the RaTE+RadGraph are all quite small compared to the gain in the Chexbert-F1-14 metric. These results confirm that a composite reward incorporating the primary target metric, F1‑14 Micro, is crucial for maximizing overall generalization.

\begin{table}[]
    \centering
    \resizebox{0.4\textwidth}{!}{
    \begin{tabular}{l|cc}
        \toprule
         \textbf{Metrics} & \textbf{RaTE-NER} & \textbf{RadGraph} \\
         \midrule
         Micro-F1-14 & \textbf{0.5120}  & 0.5009  \\
         Macro-F1-14 & \textbf{0.0662}  & 0.0556  \\
         Micro-F1-5 & \textbf{0.0839}  & 0.0432  \\
         Macro-F1-5 & \textbf{0.0502}  & 0.0215  \\
         RadGraph F1 & \textbf{0.3458} & 0.3285 \\
         RaTEScore & \textbf{0.6672}   &0.6484   \\
         \bottomrule
    \end{tabular}
    }
    \caption{Influence of editing rules on IU-XRay dataset under the SFT + EditGRPO setting.}
    \label{tab:ablation_editing}
\end{table}

\begin{table*}[ht]
\centering
\caption{Performance of various training variants on two small-scale datasets: IU-XRay and PadChest-GR.}
\label{tab:additional_results_small_scale}
\resizebox{\textwidth}{!}{%
\begin{tabular}{l|c*{7}{c}}
\toprule
\textbf{Method}
  & \textbf{Micro‐F1‑14}
  & \textbf{Macro‐F1‑14}
  & \textbf{Micro‐F1‑5}
  & \textbf{Macro‐F1‑5}
  & \textbf{RadGraph F1}
  & \textbf{RaTE}
  & \textbf{GREEN}
  & \textbf{Avg.} \\
\midrule
\rowcolor{gray!20} 
\multicolumn{9}{c}{IU-XRay} \\
\midrule
{\em Qwen2.5-VL-3B} & & & & & & & & \\
\hspace{0.5em}SFT (\textit{ep1})                           & \textbf{0.5250}  &0.0497 & 0 & 0 & 0.3001  &0.6446  &0.5806  &0.3000  \\
\hspace{0.5em}SFT(\textit{ep1}) + Dr.\ GRPO               &0.5166  &0.0613 &0.0143  &0.0053  &0.3309  &0.6508  &0.5982 &0.3111  \\
\hspace{0.5em}SFT(\textit{ep1}) + EditGRPO                   & 0.5120  & \textbf{0.0662}  &\textbf{0.0839}  &\textbf{0.0502}  &\textbf{0.3458} & \textbf{0.6672}  &\textbf{0.6517}  & \textbf{0.3396}  \\
\midrule
\rowcolor{cyan!20} 
\multicolumn{9}{c}{PadChest-GR} \\
\midrule
{\em Qwen2.5-VL-3B} & & & & & & & & \\
\hspace{0.5em}SFT (\textit{ep3})                          & 0.4533  & 0.0768  & 0.0114  & 0.0056  & \textbf{0.0844}  & 0.2541  &0.2617   & 0.1639  \\
\hspace{0.5em}SFT (\textit{ep2}) + Dr. GRPO                   &0.4128 &\textbf{0.1194} &0.0952 &0.0559 &0.0446 &0.2872  &0.2551 &  0.1815    \\
\hspace{0.5em}SFT (\textit{ep2}) + EditGRPO                  &\textbf{0.4608}  &0.1153   &\textbf{0.1017}   &\textbf{0.0708}   &0.0839   &\textbf{0.3641}   &\textbf{0.2683}     &\textbf{0.2093}  \\
\bottomrule
\end{tabular}%
}
\end{table*}

\begin{table*}[ht]
\centering
\caption{Out-of-domain generalization performance of models trained on MIMIC-CXR using various training strategies, evaluated on two small-scale datasets: IU-XRay and PadChest-GR.}
\label{tab:additional_results_generalization}
\resizebox{\textwidth}{!}{%
\begin{tabular}{l|c*{7}{c}}
\toprule
\textbf{Method}
  & \textbf{Micro‐F1‑14}
  & \textbf{Macro‐F1‑14}
  & \textbf{Micro‐F1‑5}
  & \textbf{Macro‐F1‑5}
  & \textbf{RadGraph F1}
  & \textbf{RaTE}
  & \textbf{GREEN}
  & \textbf{Avg.} \\
\midrule
\rowcolor{gray!20} 
\multicolumn{9}{c}{IU-XRay} \\
\midrule
{\em Qwen2.5-VL-3B-MIMIC-CXR} & & & & & & & & \\
\hspace{0.5em}SFT (\textit{ep3})                           & 0.5759 &0.1682 & 0.2581 &0.2050 &\textbf{0.3161} &0.6529 &0.6423 &0.4026  \\
\hspace{0.5em}SFT(\textit{ep2}) + Dr.\ GRPO               &0.5498  &0.1848 &0.3386  &0.2591  &0.2983  &0.6363 &\textbf{0.6452} &0.4160  \\
\hspace{0.5em}SFT(\textit{ep2}) + EditGRPO                   & \textbf{0.5833} &\textbf{0.2040} &\textbf{0.3644} &\textbf{0.2775} &0.3136 &\textbf{0.6557} &0.6448 &\textbf{0.4348}  \\
\midrule
\rowcolor{cyan!20} 
\multicolumn{9}{c}{PadChest-GR} \\
\midrule
{\em Qwen2.5-VL-3B-MIMIC-CXR} & & & & & & & & \\
\hspace{0.5em}SFT (\textit{ep3})                          & 0.4128 &0.1194 &0.0952 &0.0559 &0.0446 &\textbf{0.2872} &0.2041 &0.1742 \\
\hspace{0.5em}SFT (\textit{ep2}) + Dr. GRPO                   & 0.4608 &0.1153 &0.1017 &0.0708 &\textbf{0.0839} &0.2776 &0.2379 &0.1926  \\
\hspace{0.5em}SFT (\textit{ep2}) + EditGRPO                  &\textbf{0.4841} &\textbf{0.2208} &\textbf{0.3073} &\textbf{0.2563} &0.0222 &0.2858 &\textbf{0.2397} & \textbf{0.2595} \\
\bottomrule
\end{tabular}%
}
\end{table*}

\paragraph{Controllable editing is necessary for EditGRPO.} Figure~\ref{fig:val_trend} illustrates the reward gains over training steps on MIMIC-CXR. We observe that applying paragraph-level editing leads to unstable training dynamics. In contrast, sentence-level editing results in a stable training curve with consistent improvement, without signs of performance saturation. This highlights the importance of fine-grained, controllable edits for effective reinforcement learning in report generation. In contrast, applying pure GRPO results in rapid performance saturation, achieving significantly lower reward gains, approximately half, compared to the SFT + EditGRPO setting.

\paragraph{RaTE-based editing outperforms RadGraph-based editing under the EditGRPO.} Although RaTEScore only identifies presence or absence (as
abnormality versus non-abnormality or disease versus non-disease),
RadGraph identifies three presence labels (definitely present,
definitely absent, or uncertain).  In our implementation, instances marked uncertain are excluded.
However, RadGraph does not
come with an entity embedding model, so the cosine similarity
threshold reverts to exact entity matching. Results on IU-XRay, shown in Table~\ref{tab:ablation_editing}, indicate that RadGraph-based editing underperforms. 
This likely stems from reliance on exact string matching with RadGraph, which lacks canonical concept IDs or mention‑level embeddings for semantic matching, making the procedure sensitive to phrasing \citep{radgraph}.

\paragraph{Similarity threshold selection.} We set the cosine similarity threshold $\tau$ in the editing rules from the preliminary ablation study using a 3,000-case subset of the MIMIC-CXR, with evaluation on the full test set. 
As shown in the Table \ref{tab:ablation_tau}, $\tau$=0.6 yields the best performance across five of the seven metrics. CheXbert-14-Micro and Macro scores, which are directly associated with the model's ability to detect common abnormalities, show the most significant improvements. This suggests that $\tau$=0.6 effectively balances the trade-off between different editing operations. For the RaTE and RadGraph metrics, we observed that $\tau$=0.9 achieved the best performance, although the improvement was marginal. This is consistent with the nature of these metrics, particularly RadGraph, which is more sensitive to exact wording and literal entity agreement. A higher $\tau$ enforces a stricter entity-matching criterion, which slightly benefits these metrics. However, our results confirm that a balanced, sentence-level editing approach enabled by a moderate $\tau$ is crucial for maximizing overall clinical efficacy across a range of complementary metrics, as demonstrated by the superior performance on CheXbert and GREEN scores.

\begin{table}[]
    \centering
    \resizebox{0.48\textwidth}{!}{
    \begin{tabular}{l|cccc}
        \toprule
         \textbf{Metrics} & \textbf{$\tau=0$} & \textbf{$\tau=0.3$} & \textbf{$\tau=0.6$} & \textbf{$\tau=0.9$} \\
         \midrule
         RaTEScore &0.5290   &0.5333 &0.5324 &\textbf{0.5338}   \\
         RadGraph F1 &0.2720	&0.2745	&0.2726	&\textbf{0.2769} \\
         Micro-F1-14 &0.5071	&0.4995	&\textbf{0.5124}	&0.5032 \\
         Macro-F1-14 &0.3374	&0.3290	&\textbf{0.3531}	&0.3299  \\
         Micro-F1-5 &0.5477	&0.5422	&\textbf{0.5533}	&0.5482  \\
         Macro-F1-5 &0.4656	&0.4569	&\textbf{0.4712}	&0.4707  \\
         GREEN & 0.3389 &	0.3368	&\textbf{0.3455}	&0.3433  \\
         \bottomrule
    \end{tabular}
    }
    \caption{Influence of $\tau$ on MIMIC-CXR subset.}
    \label{tab:ablation_tau}
\end{table}

\begin{table*}[ht]
\small
\centering
\caption{Per\text{-}metric deltas of EditGRPO against SFT and Dr.GRPO across three datasets. Cells show absolute differences; $^{*}$ marks per\text{-}metric significance vs SFT and $^{\dagger}$ vs Dr.GRPO (two\text{-}sided tests as defined in the text). The \texttt{Avg.} row is excluded.}
\begin{tabular}{l|cc|cc|cc}
\toprule
& \multicolumn{2}{c|}{MIMIC\text{-}CXR} & \multicolumn{2}{c|}{IU\text{-}XRAY} & \multicolumn{2}{c}{PadChest\text{-}GR} \\
Metric & $\Delta$ Edit--SFT & $\Delta$ Edit--Dr & $\Delta$ Edit--SFT & $\Delta$ Edit--Dr & $\Delta$ Edit--SFT & $\Delta$ Edit--Dr \\
\midrule
RaTE                 & +0.0278$^{*}$ & +0.0252$^{\dagger}$ & +0.0226$^{*}$ & +0.0164$^{\dagger}$ & +0.1100$^{*}$ & +0.0769$^{\dagger}$ \\
CheXbert\_14\_Micro  & +0.0450$^{*}$ & +0.0124$^{\dagger}$ & -0.0130        & -0.0046             & +0.0075$^{*}$ & +0.0480$^{\dagger}$ \\
CheXbert\_14\_Macro  & +0.0330$^{*}$ & +0.0193$^{\dagger}$ & +0.0165$^{*}$  & +0.0049$^{\dagger}$ & +0.0385$^{*}$ & -0.0041             \\
CheXbert\_5\_Micro   & +0.0499$^{*}$ & +0.0074             & +0.0839$^{*}$  & +0.0696$^{\dagger}$ & +0.0903$^{*}$ & +0.0065$^{\dagger}$ \\
CheXbert\_5\_Macro   & +0.0505$^{*}$ & +0.0219$^{\dagger}$ & +0.0502$^{*}$  & +0.0449$^{\dagger}$ & +0.0652$^{*}$ & +0.0149$^{\dagger}$ \\
GREEN                & +0.0242$^{*}$ & +0.0198$^{\dagger}$ & +0.0711$^{*}$  & +0.0535$^{\dagger}$ & +0.0066       & +0.0132$^{\dagger}$ \\
RadGraph F1          & -0.0096       & -0.0118             & +0.0457$^{*}$  & +0.0149$^{\dagger}$ & -0.0005       & +0.0393$^{\dagger}$ \\
\bottomrule
\end{tabular}
\label{tab:delta_consolidated}
\end{table*}


\paragraph{Statistical significance testing.}
We assess per-metric differences $\Delta(\text{Edit--SFT})$ and $\Delta(\text{Edit--Dr.GRPO})$ using two-sided paired tests at the sample level; Table~\ref{tab:delta_consolidated} reports absolute deltas and marks significance with $^{*}$ (vs SFT) and $^{\dagger}$ (vs Dr.GRPO) for the seven primary metrics (RaTE, CheXbert-14 Micro/Macro, CheXbert-5 Micro/Macro, GREEN, RadGraph F1; \texttt{Avg.} excluded). 
On MIMIC-CXR, EditGRPO is significant on 6/7 metrics vs SFT and 5/7 vs Dr.GRPO; on IU-XRAY, 6/7 vs SFT and 6/7 vs Dr.GRPO; on PadChest-GR, 5/7 vs SFT and 6/7 vs Dr.GRPO. 
The largest and most frequent gains occur on CheXbert-5 (micro/macro) and GREEN, while RadGraph F1 shows a small non-significant drop on MIMIC-CXR and reaches significance only vs Dr.GRPO on PadChest-GR, consistent with its sensitivity to relational structure. 


\end{document}